\pdfoutput=1
\documentclass[11pt]{article}

\usepackage[final]{acl}

\usepackage{times}
\usepackage{latexsym}

\usepackage[T1]{fontenc}
\usepackage[utf8]{inputenc}

\usepackage{microtype}

\usepackage{inconsolata}

\usepackage{graphicx}

\usepackage{amsmath}
\usepackage{amssymb}
\DeclareMathOperator*{\expec}{\mathbb{E}}
\usepackage{graphicx}
\usepackage{authblk}
\usepackage{hyperref}

\usepackage{booktabs}
\usepackage{adjustbox}
\usepackage{array}
\usepackage{multirow}
\usepackage[skip=10pt]{caption} % Add this line to adjust the space between the table and the caption
\usepackage{tabularx}
\usepackage{hyperref}
\usepackage{longtable}
\usepackage{float}
\usepackage{placeins} % Import placeins package
\usepackage{soul}
\usepackage{amsmath}

\newcommand*\samethanks[1][\value{footnote}]{\footnotemark[#1]}

\title{ARES: Alternating Reinforcement Learning and Supervised Fine-Tuning for Enhanced Multi-Modal Chain-of-Thought Reasoning\\Through Diverse AI Feedback}

\author{
    \textbf{Ju-Seung Byun\thanks{Equal contribution}},
    \textbf{Jiyun Chun\samethanks},
 \textbf{Jihyung Kil},
 \textbf{Andrew Perrault}
\\
Department of Computer Science and Engineering\\
The Ohio State University\\
\texttt{\{byun.83,chun.203,kil.5,perrault.17\}@osu.edu}\\
\\
}

\begin{document}

\maketitle

% \begingroup
% \renewcommand\thefootnote{\textsuperscript{*}}
% \footnotetext{Equal contribution}
% \endgroup

\begin{abstract}
Large Multimodal Models (LMMs) excel at comprehending human instructions and demonstrate remarkable results across a broad spectrum of tasks. Reinforcement Learning from Human Feedback (RLHF) and AI Feedback (RLAIF) further refine LLMs by aligning them with specific preferences. These methods primarily use ranking-based feedback for entire generations. With advanced AI models (Teacher), such as \verb|GPT-4| and \verb|Claude 3 Opus|, we can request various types of detailed feedback that are expensive for humans to provide. We propose a two-stage algorithm \textbf{ARES} that \textbf{A}lternates \textbf{RE}inforcement Learning (RL) and \textbf{S}upervised Fine-Tuning (SFT). First, we ask the Teacher to score how much each sentence contributes to solving the problem in a Chain-of-Thought (CoT). This sentence-level feedback allows us to consider individual valuable segments, providing more granular rewards for the RL procedure. Second, we ask the Teacher to correct wrong reasoning after the RL stage. The RL procedure requires substantial hyperparameter tuning and often generates errors such as repetitive words and incomplete sentences. With correction feedback, we stabilize the RL fine-tuned model through SFT. We conduct experiments on the multi-modal datasets ScienceQA and A-OKVQA to demonstrate the effectiveness of our proposal. The ARES rationale achieves around 70\% win rate compared to baseline models judged by \verb|GPT-4o|. Additionally, we observe that the improved rationale reasoning leads to a 2.5\% increase in inference answer accuracy on average for the multi-modal datasets. \footnote{Code: \url{https://github.com/Amyyyyeah/ARES}}
% \footnote{Our code is available in the supplementary for review material and will be publicly available.}
\end{abstract}

\section{Introduction}
Large Language Models (LLMs) and Large Multimodal Models demonstrate remarkable performance across diverse language and multi-modal tasks \citep{brown2020language, chowdhery2022palm, touvron2023llama, zhang2022opt, liu2023visual, goyal2023news}. However, these Large Models (LMs) often generate toxic and biased content \citep{gehman2020realtoxicityprompts, tamkin2021understanding} because LMs are primarily trained to predict the next token based on extensive corpus datasets. To align LM behavior more closely with user preferences, previous works \citep{glaese2022improving, ouyang2022training} fine-tune their models using Reinforcement Learning from Human Feedback (RLHF). Furthermore, with advancements in LMs, advanced LM feedback can replace costly human feedback, yielding Reinforcement Learning from AI Feedback (RLAIF) \citep{lee2023rlaif, bai2022constitutional, yuan2024selfrewarding}.
\begin{figure}[t]
  \includegraphics[width=\columnwidth]{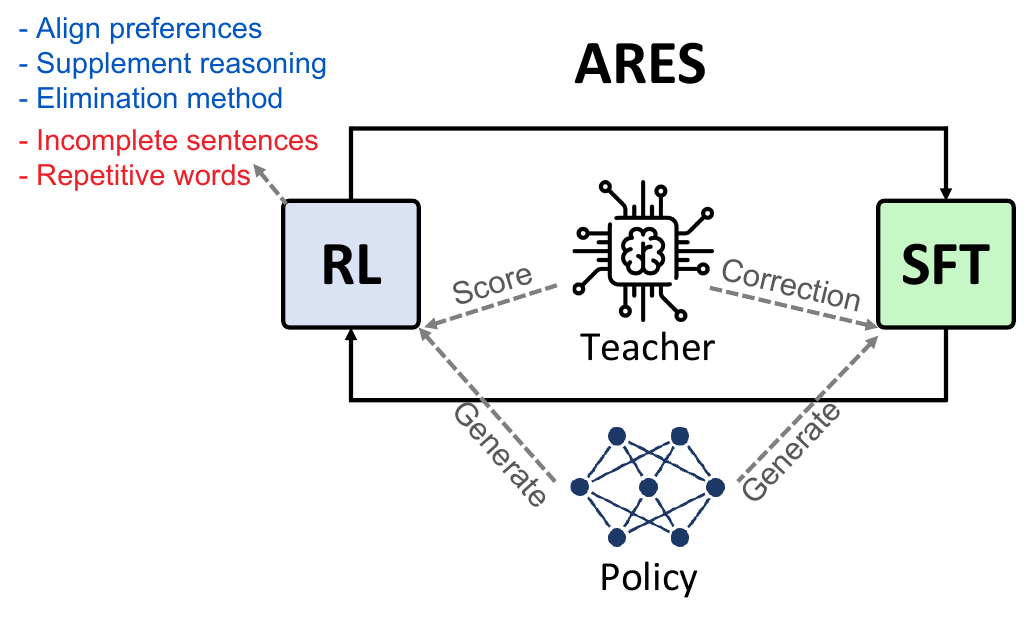}
  \vspace{-1.8em}
  \caption{\textbf{Overview of ARES:} ARES alternates RL and SFT. The Teacher provides scores (rewards) for each sentence for RL. \textcolor{blue}{Blue} indicates the advantages of RL, and \textcolor{red}{red} indicates potential degeneration. ARES corrects the issues through the Teacher's correction feedback.}
  \label{ares_simple_pipeline}
  \vspace{-1.0em}
\end{figure}

However, RLHF and RLAIF encounter two significant challenges. First, both methods often utilize ranking-based feedback~\citep{ouyang2022training}, which orders the generated samples by preference. For instance, if sample $A$ is preferred over sample $B$, the model is fine-tuned to generate more outputs like $A$ and fewer like $B$. However, if $B$ contains certain valuable parts, they may be discarded. To alleviate these issues, \citet{lightman2023lets, luo2024improve} propose sentence-level feedback, applying it solely to the reward model without Reinforcement Learning (RL), called the Process-supervised Reward Model (PRM). It demonstrates its potential through search algorithms on the PRM, such as best-of-$N$ or Monte Carlo Tree Search (MCTS). Furthermore, \citet{wang2024mathshepherd} demonstrate the effectiveness of RL with sentence-level feedback by heuristically scoring each sentence in math problems, where evaluating the predicted answer is straightforward. Thus, sentence-level feedback exhibits significant promise compared to existing ranking-based feedback. Nonetheless, acquiring sentence-level feedback is more costly than ranking-based feedback.

Second, the RL process is inherently unstable and requires extensive hyperparameter tuning \citep{eimer2023hyperparameters}. This instability often results in the generation of repetitive words and truncated sentences. Hyperparameter tuning becomes an enormous burden as the model size increases, making exhaustive tuning seemingly impossible, especially for individuals. The existing RLHF method \citep{ouyang2022training} recycles the dataset used in pretraining within the loss function with Proximal Policy Optimization (PPO) \citep{DBLP:journals/corr/SchulmanWDRK17} to mitigate this degeneration problem. However, this approach prevents the model from fully maximizing the sum of rewards through RL and may limit the opportunity for diverse improvements, which differ from the pretraining distribution.

In this work, we aim to address the two challenges mentioned above through various types of feedback using an advanced AI model as a Teacher. Many advanced AI models, including \verb|GPT-4| and \verb|Claude 3 Opus|, are already used as evaluators for many tasks and generate reliable human-level answers \citep{liu2023geval, sottana2023evaluation}. $1)$ We request a score from the Teacher for each sentence, ranging from $0.0$ to $1.0$. Each score indicates how much a sentence contributes to solving the problem. This provides detailed reward feedback to the training model and can be applied to both mathematical and more general multi-modal Chain-of-Thought (CoT) (or rationale reasoning) problems. $2)$ We ask the Teacher to identify and correct minor errors in the RL results, such as incorrect or cut-off parts. With this corrected dataset, we fine-tune the model using Supervised Fine-Tuning (SFT). This stage allows the model to maximize the rewards while properly deviating from the pretraining distribution. In summary, we propose a hybrid algorithm \textbf{ARES} that \textbf{A}lternates \textbf{RE}inforcement Learning and \textbf{S}upervised Fine-Tuning.

To evaluate how much rationale reasoning can be improved through the ARES framework, we use the ScienceQA and A-OKVQA datasets, which are large-scale, multi-modal datasets that include rationale reasoning data. We use Multimodal-CoT (MM-CoT) \citep{zhang2023multimodal} as the baseline. MM-CoT employs two separate models: one model is responsible for generating rationale, and the other model, an inference model, processes the concatenated input (problem and generated rationale). This distinct framework  enhances performance, even with relatively smaller models like $\text{Flan-Alpaca}_{\text{Base}}$ \citep{chia2023flanalpaca} (251M) and $\text{Flan-Alpaca}_{\text{Large}}$ (790M) with ViT feature \citep{dosovitskiy2021image}. We perform ARES on the rationale reasoning model of MM-CoT. We compare ARES rationale reasoning with that of MM-CoT through \verb|GPT-4o|, determining which rationale is better and computing the win rate. Additionally, we check whether the improved rationale reasoning leads to better answer accuracy. Our results show that our rationale reasoning outperforms the baselines with around 70\% win rate and demonstrates 2.5\% increase in inference answer accuracy on average for the different model sizes and the multi-modal tasks. 

In summary, our key contributions are:
\begin{itemize}
\item We propose ARES, leveraging diverse types of feedback from advanced AI (Teacher) to enhance CoT reasoning for multi-modal tasks.
\item ARES properly reflects the direction in which the model is fine-tuned through RL and stabilizes the RL fine-tuning procedure with SFT.
\item ARES generates better rationale chains compared to baselines judged by \verb|GPT-4o| and improves inference answer accuracy for multi-modal tasks.
\end{itemize}

\section{Methodology}
This section briefly introduces the preliminaries in Section~\ref{rl_background} and presents our two-stage hybrid algorithm \textbf{ARES} that \textbf{A}lternates \textbf{RE}inforcement Learning and \textbf{S}upervised Fine-Tuning. $1)$ We request a score for each sentence in the Chain-of-Thought (CoT) from the advanced AI model (Teacher) to determine how much it contributes to solving the problem (Section~\ref{rl_stage}). We perform Reinforcement Learning (RL) with the score feedback on our training model. $2)$ The Teacher corrects minor errors such as truncated or slightly incorrect sentences, thereby performing Supervised Fine-Tuning (SFT) (Section~\ref{rl_stage}).

\subsection{Preliminaries}
\label{rl_background}
For an input $x \in X$, a Transformer-based \citep{vaswani2023attention} model $\pi_{\theta}(\cdot|x)$ parameterized by $\theta$ generates the output $y$ composed of sentences $\{s_0, s_1,...,s_k\}$. 
\begin{equation}
\pi_{\theta}(y \mid x) = \prod_{t=0}^{k} \pi_{\theta}(y_i \mid x, y_{<t}),
\end{equation}
where $y_{<t}$ indicates previous tokens. To proceed with RL finetuning, \citet{ouyang2022training} train an outcome-supervised reward model (ORM) using ranking-based feedback. With more fine-grained feedback like sentence-level, \citet{lightman2023lets, wang2024mathshepherd} train a process-supervised reward model (PRM). Instead of training a reward model, we request score feedback $r(x \cup s_{<t}, s_t)$ for each sentence from an advanced AI such as \verb|GPT-4| where $s_{<t}$ represents previous sentences. 

\begin{figure*}[t]
\label{reward_model_error}
\centering
\includegraphics[scale=0.52]{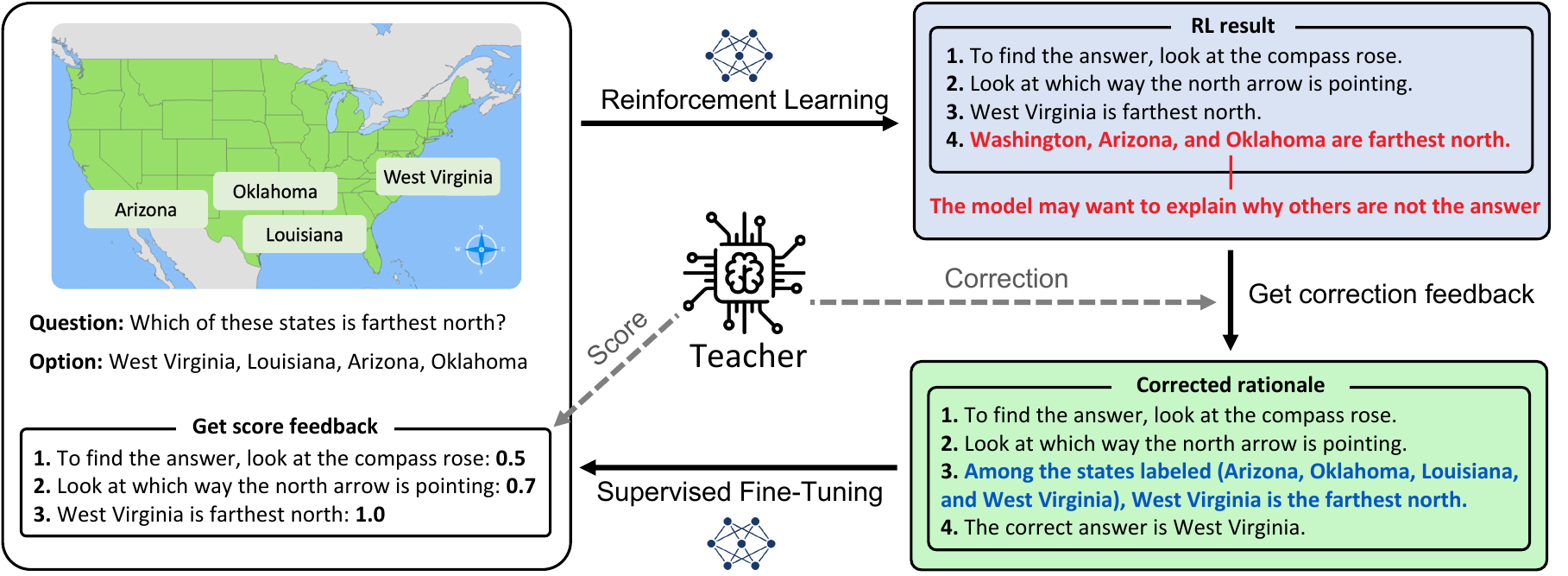}
\caption{ARES pipeline: For a given model generating rationale reasoning, we request an AI model's sentence-level scores (ranging from $0.0$ to $1.0$). The closer the score is to $1.0$, the more it helps solve the problem. We proceed with the RL stage using these sentence-level scores. After RL, the training model may produce incorrect parts (colored in red), so we enhance the training model by requesting correction feedback from the AI model (colored in blue) and then proceed with the supervised fine-tuning stage.}
\label{rl_sft}
\end{figure*}

\subsection{Reinforcement Learning}
\label{rl_stage}
Reinforcement Learning (RL) fine-tunes our model $\pi_{\theta}$ to maximize sum of sentence rewards from an advanced AI model such as \verb|GPT-4| and \verb|Claude 3 Opus|. The RL objective is as follows:
\begin{align}
\label{rl_objective}
\theta^{*} = \underset{\theta}{\mathrm{argmax}}  \expec_{\substack{x \sim X \\ s_t \sim \pi_{\theta}(\cdot | x, s_{<t})}}\Biggl[\sum_{i = 0}^{k}\gamma^{i} r(x \cup s_{<t}, s_t)\Bigg ].
\end{align}
where $\gamma \le 1.0$ is a discount factor.  We use Proximal Policy Optimization (PPO) \citep{DBLP:journals/corr/SchulmanWDRK17} to achieve this RL objective, treating sentences as actions (Equation~\ref{ppo_eq}).
\begin{equation}\begin{aligned}
    & \underset{\theta}{\rm{maximize}}\; \mathbb{E}_{t}\bigg[\frac{\pi_{\theta}(s_{t}|x \cup s_{<t})}{\pi_{\theta_{\rm{old}}}(s_{t}|x \cup s_{<t})}\hat A_t  \bigg] \\
    & \text{s.t.}\; \mathbb{E}_{t}\big[ \rm{KL}\left(\pi_{\theta}(\cdot|x \cup s_{<t}),\pi_{\theta_{old}}(\cdot|x \cup s_{<t})\right)\big] \leq \delta
\end{aligned}\label{ppo_eq}\end{equation}
where $\pi_{\rm{old}}$ is the original policy (baseline model) and $\hat A_t$ is an advantage estimator at timestep $t$. PPO is commonly leveraged in Reinforcement Learning from Human Feedback (RLHF) \citep{ouyang2022training} and AI Feedback (RLAIF) \citep{bai2022constitutional}. PPO's conservative update prevents the training model from deviating too far from the original model, thus avoiding degeneration.\\

\noindent\textbf{Sentence-Level Nuanced Feedback:} We request a score between $0.0$ and $1.0$ for each sentence in CoT through the advanced AI for RL. The closer the score is to $1.0$, the more relevant and helpful it is to solving the problem. Table~\ref{fig:sentence_prompt} presents the prompt format. We additionally shift the reward distribution by $-0.5$ to center it at $0$ \citep{zheng2023secrets}. Therefore, the actual range is from $-0.5$ to $0.5$. Using these nuanced scores, the RL fine-tuned model  exhibits emergent behaviors (please refer to Section~\ref{experimental_results}). This allows us to understand the direction in which the model is intended to change through RL. \\

\noindent\textbf{Advantages of Using Advanced AI for Score Feedback:} Although calling the API has disadvantages, such as incurring costs or facing usage limits, there exist several advantages to using the advanced AI for feedback. First, there is no need to train a reward model. Second, as the RL fine-tuned model begins to generate out-of-distribution outputs that differ from the data used to train the reward model, it becomes challenging for the trained reward model to provide accurate rewards. However, this out-of-distribution problem is effectively addressed with the advanced AI. \\

\noindent\textbf{RL Challenge:} 
\label{sec:RLchallenges}
One of the challenging factors for RL is hyperparameter tuning \citep{eimer2023hyperparameters}. This often results in generating repetitive words and truncated sentences \citep{ouyang2022training}. Additionally, as the model size increases, finding working hyperparameters becomes infeasible for individuals. To alleviate this issue, we utilize correction feedback from the advanced AI as the second stage (Section~\ref{sl_stage}), and proceed with the supervised fine-tuning to stabilize the RL fine-tuned model. 

\subsection{Correction: Supervised Fine-Tuning}
\label{sl_stage}
The RL fine-tuning procedure makes model changes to maximize the reward sum, such as correcting mistakes or explaining why other options cannot be the answer. However, without highly tuned hyperparameters \citep{eimer2023hyperparameters}, the model after the RL phase may result in errors such as repeated sentences, truncated sentences, or incorrect content for some data points. (See examples in Appendix~\ref{appendix_comparision}.) \\

\noindent\textbf{Correction Feedback:} Given the success of LLMs and LMMs in a wide range of areas \citep{brown2020language, chowdhery2022palm, zhang2022opt}, we are not restricted to requesting feedback in the form of scores. We request correction feedback from advanced AI (Teacher) for sentences containing errors after the RL process, and obtain a corrected dataset $X_{\text{corrected}}$. Since the supervised fine-tuning is more stable and finding appropriate hyperparameters is easier than RL, we proceed with supervised fine-tuning using $X_{\text{corrected}}$ exactly as in common autoregressive model \citep{vaswani2023attention} training to stabilize the RL fine-tuned model. This reduces the burden of RL’s exhaustive hyperparameter tuning and properly guides the direction in which the training model wants to change. \\

\noindent\textbf{How Correction Feedback Helps RL:} RL increases the probability of positively rewarded actions (or sentences) and decreases the probability for negative rewards. The direction of learning is determined by the reward (scalar) value. However, the opposite direction of the reward is sometimes required. For example, suppose there is a truncated sentence $s_{\text{truncated}}$ in CoT. $s_{\text{truncated}}$ gets a negative score because it is an incomplete sentence (Table~\ref{table-incompete}). If there is no correction stage, the probability of $s_{\text{truncated}}$ is simply reduced. \textit{What if $s_{\text{truncated}}$ contains some valuable part?} This valuable part is ignored, and its probability decreases. To alleviate this issue, we instead receive the corrected sentence as feedback and encourage the training model to generate complete sentences, which is very challenging to achieve with only RL. Table~\ref{fig:correction_example} shows more examples of how the correction stage helps the RL stage by maintaining the reasoning context while changing the erroneous parts.

Additionally, RL is primarily fine-tuned through PPO \citep{DBLP:journals/corr/SchulmanWDRK17} to prevent the model from deviating too much from the original model. The KL divergence penalty further prevents deviation. However, this penalty often causes the model's degeneration. As a solution, InstructGPT \citep{ouyang2022training} proposes PPO-ptx, where the supervised fine-tuning term with the pretraining dataset is included in the loss function. While this aims to align the training model with specific preferences, it tends to anchor the model to the pretraining dataset. Instead, we perform supervised fine-tuning through the Teacher's correction feedback to allow the training model to more freely  adapt and meet specific preferences without degeneration.

\subsection{Algorithm Details}
We propose a hybrid algorithm \textbf{A}lternating between \textbf{RE}inforcement learning and \textbf{S}upervised fine-tuning (ARES). Figure~\ref{rl_sft} illustrates the ARES pipeline. First, we prepare a model with a given training dataset and generate rationale reasoning composed of several sentences for input. For the RL procedure to align the training model with a preference, we request scores for each sentence. The RL result may include some incorrect parts (colored in red as the 4th sentence in the RL result box), but it aims to maximize the rewards provided. Next, we request correction feedback and create a corrected dataset (colored in blue as the 3rd sentence in the Corrected Rationale box) for the supervised fine-tuning stage. We then repeat the process from the RL stage until convergence.

\section{Experimental Setup}
\noindent\textbf{Data:} We first evaluate our proposed method on the ScienceQA \citep{lu2022learn} dataset, a large-scale, multi-modal science dataset designed to assess multi-hop reasoning abilities. We choose ScienceQA because it contains reasoning chains to derive the answer. Each problem consists of a question, multiple options, multi-modal contexts, a correct answer, and an annotated lecture or solution chain (note that around $9.5\%$ lack the solution chain). %ScienceQA has $21$K multi-modal problems, with $12$K for training, $4$K for validation, and $4$K for testing. It also includes various difficulty levels from elementary to high school, covering domains like natural science, language science, and social science. 
In addition, we conduct experiments on A-OKVQA \citep{schwenk2022aokvqa}, a knowledge-based multi-modal benchmark with a diverse set of challenging questions paired with rationales, demanding non-trivial commonsense knowledge (see Appendix~\ref{appendix_vqa_challenge}). \\

%In addition, we conduct experiments on A-OKVQA \citep{schwenk2022aokvqa}, a knowledge-based multi-modal benchmark with a diverse set of 25K questions A-OKVQA includes 25K questions ($17$K for training, $1$K for validation, and $6$K for testing). A-OKVQA includes challenging questions paired with rationales, demanding non-trivial commonsense knowledge (see Appendix~\ref{appendix_vqa_challenge}). \\

\noindent\textbf{Baselines:} We mainly compare our method with Multimodal-CoT (MM-CoT) \citep{zhang2023multimodal} as the baseline because it utilizes reasoning chains to solve multi-modal tasks. MM-CoT leverages two distinct models: the first generates a rationale for a given problem, and the second, an inference model, takes the concatenated input (problem and generated rationale). This separated framework shows improved performance, even for relatively small models such as $\text{Flan-Alpaca}_{\text{Base}}$ \citep{chia2023flanalpaca} ($251$M) and $\text{Flan-Alpaca}_{\text{Large}}$ ($790$M). We use the rationale model provided by MM-CoT for ScienceQA and retrain the rationale model ourselves for A-OKVQA because there is no provided model.\\

\noindent\textbf{Prompts for Feedback:} Since our proposed ARES requests different types of feedback for each stage, a corresponding prompt exists separately. We use \verb|Claude 3 Haiku| for all training to get feedback because it is approximately $20$ times cheaper than the top competing models, yet still demonstrates decent performance. We first request scores ranging from $0.0$ to $1.0$ for each sentence in CoT to proceed with the RL stage. To obtain reasonable scores, we let \verb|Haiku| consider the starting point of thought, the process of elimination, or true statements. (See Table~\ref{fig:sentence_prompt}.)

In order to collect the corrected dataset for the SFT stage, we let \verb|Haiku| refer to the given problem and correct the answer as the prompt. We ask \verb|Haiku| to maintain the format of the existing rationale chains as much as possible and correct only the parts that require correction. The RL stage often makes the training model generate repetitive sentences. This repetition is not easily removed even by \verb|GPT-4| when the repetitive sentence exists in the middle of rationale reasoning. To reduce the burden of feedback, we simply hard-code the removal of repetitive sentences before adding the generated rationale to the prompt. (See Appendix~\ref{appendix_sft}.)\\

% \noindent\textbf{Training Details:} For the $\text{ARES}_{\rm{Base}}$ RL stage, we use a learning rate of $2e{-5}$ and $10$ epochs for PPO with a batch size of $8$ for both ScienceQA and A-OKVQA. The learning rate for $\text{ARES}_{\rm{Large}}$ is $2e{-5}$ with $5$ epochs for PPO and a batch size of $2$ for both tasks. We proceed with 2 rounds of our pipeline for $\text{ARES}_{\rm{Base}}$ and 2 rounds for $\text{ARES}_{\rm{Large}}$ for ScienceQA. For A-OKVQA, we proceed with 1 round for both model sizes. For the SFT stage for correction, we follow the hyperparameters used in MM-CoT for both model sizes. Additionally, we replace MM-CoT's inference model, which is the same size as the rationale model, with a LoRA adapter added to the rationale model (Figure~\ref{fig:lora}). The LoRA adapter effectively refers to the rationale model's features with a small number of weights and infers answers . Please refer to the more detailed settings in Appendix~\ref{appendix_training_details}.\\

\noindent\textbf{Training Details:} For the $\text{ARES}_{\rm{Base}}$ RL stage, we use a learning rate of $2e{-5}$ and $10$ epochs for PPO with a batch size of $8$ for both ScienceQA and A-OKVQA. The learning rate for $\text{ARES}_{\rm{Large}}$ is $2e{-5}$ with $5$ epochs for PPO and a batch size of $2$ for both tasks. We proceed with 2 rounds of our pipeline for $\text{ARES}_{\rm{Base}}$ and 2 rounds for $\text{ARES}_{\rm{Large}}$ for ScienceQA. For A-OKVQA, we proceed with 1 round for both model sizes. For the SFT stage for correction, we follow the hyperparameters used in MM-CoT for both model sizes. Additionally, we replace MM-CoT's inference model, which is the same size as the rationale model, with the Low-Rank Adaptation (LoRA) \citep{hu2021lora} added to the rationale model (Figure~\ref{fig:lora}). The LoRA adapter effectively utilizes the rationale model's features with a small number of weights, enabling 2x--14x faster inference compared to MM-CoT, which introduces a separate inference model (See the time comparison in Table~\ref{time_comparison_scienceQA} and Table~\ref{time_comparison_aokvqa}). For more detailed settings, please refer to Appendix~\ref{appendix_training_details}.\\

\noindent\textbf{Evaluation Metrics:} We use two main metrics to test how our pipeline (ARES) improves rationale reasoning quality. First, we evaluate ARES's rationale reasoning quality against baseline models since we enhance our model based on them. For two different model sizes ($\text{Flan-Alpaca}_{\rm{Base}}$ and $\text{Flan-Alpaca}_{\rm{Large}}$) and two tasks (ScienceQA and A-OKVQA), rationale reasoning quality is evaluated by \verb|GPT-4o-2024-05-13| and the win rate is calculated (Section~\ref{ex_results_win_rate}). The \verb|GPT-4| series is actively used as an evaluation metric, replacing human judgment for various domains \citep{liu2023geval, sottana2023evaluation}. Second, we assess how the improved rationale reasoning impacts answer accuracy (Section~\ref{ex_results_accuracy}). This evaluation is also performed on both model sizes and tasks. Additionally, we analyze how the RL stage fine-tunes the training model and maximizes the sum of rewards in Section~\ref{ex_results_rl}.

\section{Experimental Results}
\label{experimental_results} 
In this section, we evaluate our pipeline ARES that \textbf{A}lternates \textbf{RE}inforcement Learning and \textbf{S}upervised Fine-Tuning by requesting diverse types of feedback for an advanced AI model (Teacher) (\verb|Claude 3 Haiku|). The goal of ARES is to improve the rationale reasoning quality. We demonstrate how ARES enhances rationale reasoning in the following sections.

\subsection{Emergent Behavior Through RL} 
\label{ex_results_rl}
Through RL, a training model is aligned to a specific preference. Essentially, the model increases the probability of helpful sentences receiving good rewards and reduces the probability of incorrect or meaningless sentences. However, this process produces some interesting additional results.

First, it supplements rationale reasoning for some problems where rationale reasoning is insufficient. In particular, 9.5\% of problems in ScienceQA have empty rationale reasoning (solution) data. The model generates nothing before the RL stage for these problems but starts generating reasoning chains afterward (See Table~\ref{table-empty-filling}). We observe this especially when utilizing PPO's advantage normalization or when the learning rate is large.

Second, the training model begins to explain why other options are not the answer (See Table~\ref{table-process-elimination}). The process of elimination is a useful method for deriving answers when options are given.

\subsection{Guide RL with Correction} 
\label{ex_results_correction}
Despite the benefits of RL, hyperparameter tuning often requires massive effort. Without meticulous tuning, the RL fine-tuned model may produce errors such as repetitive or incomplete sentences. To address these issues, we add a supervised fine-tuning (SFT) stage after RL to correct these errors. SFT is more stable than RL. We evaluate how well the SFT stage corrects errors caused by the RL stage for various RL hyperparameters. We test various RL hyperparameters such as learning rate = \{5e-6, 1e-5, 2e-5, 5e-5\}, batch size = \{2, 4, 8, 16, 32\}, and PPO epoch = \{5, 10, 15\}. As a result of RL, we observe that some of the sentences in rationale chains are repetitive or truncated (see Table~\ref{table-incompete} and~\ref{table-repeat}). The SFT stage, with correction feedback, reflects the direction in which the model is fine-tuned through RL and appropriately guides it (Table~\ref{table-incompete} and~\ref{fig:correction_example}).  However, excessive RL learning rates or epochs cause serious degeneration of the model, such as producing no output or generating strange words, and the results of correction feedback are also unreasonable. 
\begin{table}[t]
\centering
\begin{adjustbox}{max width=\textwidth}
\begin{tabular}{lc}
\toprule
\textbf{ScienceQA} & \textbf{Win Rate} \\
\midrule
$\textbf{ARES}_{\rm{Base}}$ vs $\text{MM-CoT}_{\rm{Base}}$  & 69.76\% \\
$\textbf{ARES}_{\rm{Large}}$ vs $\text{MM-CoT}_{\rm{Large}}$ & 73.76\% \\
\midrule
\midrule
\textbf{A-OKVQA} & \textbf{Win Rate} \\
\midrule
$\textbf{ARES}_{\rm{Base}}$ vs $\text{MM-CoT}_{\rm{Base}}$ & 69.11\% \\
$\textbf{ARES}_{\rm{Large}}$ vs $\text{MM-CoT}_{\rm{Large}}$ & 66.96\% \\
\bottomrule
\end{tabular}
\end{adjustbox}
\caption{We train baseline models, MM-CoT, with the ARES pipeline and ask \texttt{GPT-4o} to evaluate which rationale reasoning is better. We compare each baseline for two model sizes ($\textbf{ARES}_{\rm{Base}}$ and $\textbf{ARES}_{\rm{Large}}$) and two tasks (ScienceQA and A-OKVQA).}
\label{table-win-rate}
\end{table}
\begin{table*}[ht]
\centering
%\caption{Performance of various models on different tasks.}
\begin{adjustbox}{max width=\textwidth}
\begin{tabular}{lcccccccccc}
\toprule
\textbf{Model} & \textbf{Size} & \textbf{NAT} & \textbf{SOC} & \textbf{LAN} & \textbf{TXT} & \textbf{IMG} & \textbf{NO} & \textbf{G1-6} & \textbf{G7-12} & \textbf{Avg} \\
\midrule
Human & - & 90.23 & 84.97 & 87.48 & 89.60 & 87.50 & 88.10 & 91.59 & 82.42 & 88.40 \\
\midrule
MCAN \citep{yu2019deep} & 95M & 56.08 & 46.23 & 58.09 & 59.43 & 51.17 & 55.40 & 51.65 & 59.72 & 54.54 \\
Top-Down \citep{anderson2018bottomup} & 70M & 59.50 & 54.33 & 61.82 & 62.90 & 54.88 & 59.79 & 57.27 & 62.16 & 59.02 \\
BAN \citep{kim2018bilinear} & 112M & 60.88 & 46.57 & 66.64 & 62.61 & 52.60 & 65.51 & 56.83 & 63.94 & 59.37 \\
DFAF \citep{peng2019dynamic} & 74M & 64.03 & 48.82 & 63.55 & 65.88 & 54.49 & 64.11 & 57.12 & 67.17 & 60.72 \\
ViLT \citep{kim2021vilt} & 113M & 60.48 & 63.89 & 60.27 & 63.20 & 61.38 & 57.00 & 60.72 & 61.90 & 61.14 \\
Patch-TRM \citep{lu2022iconqa} & 90M & 65.19 & 46.79 & 65.55 & 66.96 & 55.28 & 64.95 & 58.04 & 67.50 & 61.42 \\
VisualBERT \citep{li2019visualbert} & 111M & 59.33 & 69.18 & 61.18 & 62.71 & 62.17 & 58.54 & 62.96 & 59.92 & 61.87 \\
\midrule
UnifiedQA\textsubscript{Base} \citep{khashabi2020unifiedqa} & 223M & 68.16 & 69.18 & 74.91 & 63.78 & 61.38 & 77.84 & 72.98 & 65.00 & 70.12 \\
UnifiedQA\textsubscript{Base} w/ CoT \citep{lu2022learn} & 223M & 71.00 & 76.04 & 78.91 & 66.42 & 66.53 & 81.81 & 77.06 & 68.82 & 74.11 \\
\midrule
LLaMA-Adapter \citep{zhang2023llamaadapter} & 6B & 84.37 & 88.30 & 84.36 & 83.72 & 80.32 & 86.90 & 85.83 & 84.05 & 85.19 \\
LLaVA \citep{liu2023visual} & 13B & 90.36 & 95.95* & 88.00 & 89.49 & 88.00 & 90.66 & 90.93 & 90.90* & 90.92 \\
InstructBLIP \citep{dai2023instructblip} & 11B & - & - & - & - & 90.70* & - & - & - & - \\
\midrule
$\text{MM-CoT}_{\rm{Base}}$ \citep{zhang2023multimodal} & 251M+251M & 84.59 & 92.46 & 83.45 & 83.87 & 83.29 & 85.64 & 86.34 & 85.23 & 85.95 \\
$\textbf{ARES}_{\rm{Base}}$ (Ours)  & 251M+30M & \textbf{87.92} & \textbf{92.58} & \textbf{85.91} & \textbf{86.61} & \textbf{85.82} & \textbf{88.36} & \textbf{88.88} & \textbf{87.48} & \textbf{88.38}\\
\midrule
$\text{MM-CoT}_{\rm{Large}}$ \citep{zhang2023multimodal} & 790M+790M & 90.76 & 93.59 & 86.55 & 89.69 & 87.85 & 89.55 & 90.90 & 89.12 &90.26  \\
$\textbf{ARES}_{\rm{Large}}$ (Ours) & 790M+76M & \textbf{91.21*} & 92.80 & \textbf{89.45*} & \textbf{90.27*} & \textbf{88.35} & \textbf{91.22*} & \textbf{91.48*} & \textbf{90.38} & \textbf{91.09*} \\
\bottomrule
\end{tabular}
\end{adjustbox}
\centering
\caption{\label{table-scienceqa}
    Main results on the ScienceQA test set (\%). Size = backbone size. Question classes: NAT = natural science, SOC = social science, LAN = language science, TXT = text context, IMG = image context, NO = no context, G1-6 = grades 1-6, G7-12 = grades 7-12. Other results are sourced from \citet{lu2022learn} and \citet{zhang2023multimodal}. Results in bold represent the better performance corresponding baseline. (*) indicates the best performance. 
  }
  % \vspace{-1.0em}
\end{table*}

\subsection{Rationale Reasoning Comparison} 
\label{ex_results_win_rate}
We check whether ARES improves the quality of rationale reasoning compared to the baseline model. \verb|GPT-4o| evaluates which rationale chain is better between the rationale generated by ARES and the rationale generated by the baseline model. We randomly shuffle the rationale chains and provide them as Option A and Option B (see Appendix~\ref{appendix_win_rate}) for a fair evaluation \citep{yu2023promptbased}. We conduct our experiments with two different model sizes, Flan-Base and Flan-Large with ViT feature, on ScienceQA and A-OKVQA. Table~\ref{table-win-rate} shows that ARES achieves around 70\% win rate against each corresponding baseline model for both datasets.

\subsection{Inference Accuracy}
\label{ex_results_accuracy}
We investigate whether the improved rationale also contributes to answer inference accuracy. Table~\ref{table-scienceqa} shows the main results of answer inference on the ScienceQA. We evaluate our base model against the MM-CoT baseline. $\textbf{ARES}_{\rm{Base}}$ achieves a 2.79\% improvement compared to the corresponding baseline ($\text{MM-CoT}_{\text{Base}}$). The large model ($\textbf{ARES}_{\rm{Large}}$) shows some minimal improvement compared to the corresponding baseline. However, it's worth noting that despite this seemingly small gain, $\textbf{ARES}_{\rm{Large}}$ beats $13$B LLaVA \citep{liu2023visual}. This minimal improvement may be due to the 9.5\% of ScienceQA problems needing more rationale reasoning (around 9.5\% problems have empty rationale reasoning). The RL stages can only eliminate some empty rationale reasoning, which requires numerous ARES pipeline rounds. Above all, our main goal is to assess how the RL stage works and how the SFT stage aids RL.

Table~\ref{table-a-okvqa} shows the results of answer inference on the A-OKVQA. We retrain $\text{MM-CoT}_{\text{Base}}$ and $\text{MM-CoT}_{\text{Large}}$ and evaluate these on the validation set as in \citep{zhang2023multimodal} because the test set is hidden. In our experiments, MM-CoT models perform around 10\% better than the reported accuracy in \citep{zhang2023multimodal}. ARES achieves 4.45\% gains against $\text{MM-CoT}_{\text{Base}}$ and 2.35\% for $\text{MM-CoT}_{\text{Large}}$.

\begin{figure}[t]
  \includegraphics[width=\columnwidth]{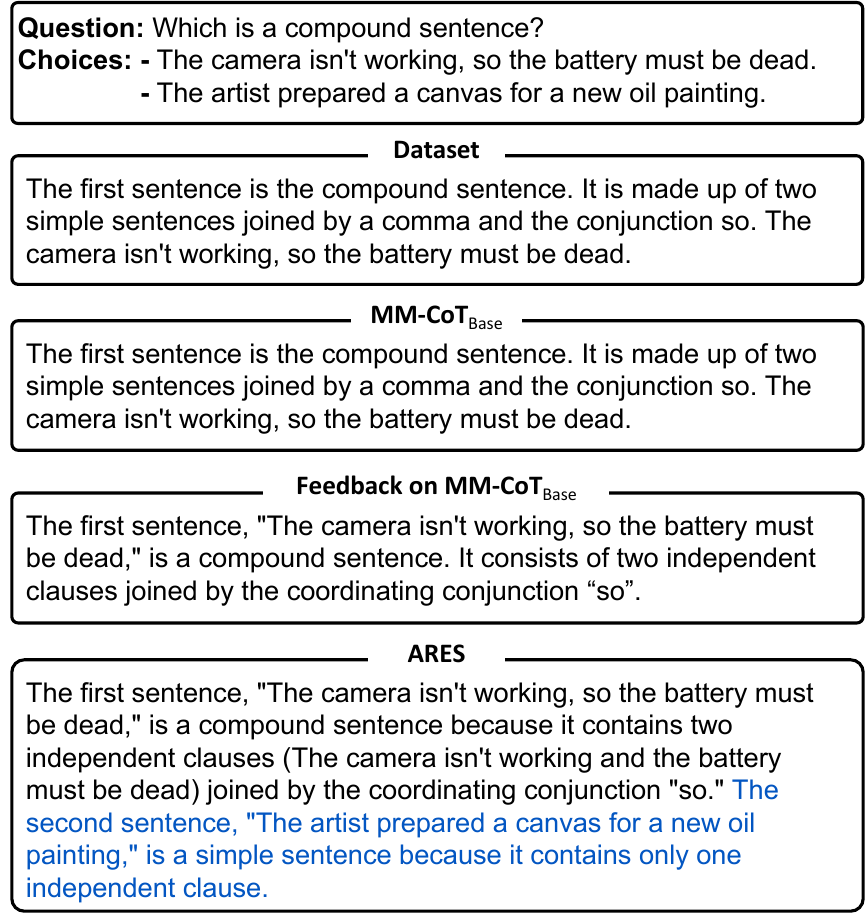}
   \caption{Comparison of rationales: dataset, baseline, correction feedback for baseline, and ARES.}
  \label{fig:main_good_example}
\end{figure}

In addition, we demonstrate that two stages, RL and SFT, are essential through an ablation study. Figure~\ref{fig:main_good_example} shows the rationale reasoning for 4 cases. The baseline model (MM-CoT) produces the same rationale reasoning as the dataset. However, the corrected reasoning for MM-CoT without the RL stage has insufficient information compared to the reasoning of ARES that performs RL (refer to Table~\ref{fig:good_example} for more examples). Table~\ref{table-ablation} also shows that inference accuracy gradually improves as each part of ARES is executed. $\text{1st}$ RL indicates a single RL run on MM-CoT, and $\text{1st ARES}$ means one round of the ARES pipeline. $\text{1st ARES \& 2nd RL}$ represents the second RL on $\text{1st ARES}$, and finally,  \text{2nd ARES} refers to two rounds of ARES.

\begin{table}[ht]
\centering
\begin{adjustbox}{max width=\textwidth}
\begin{tabular}{lcccccccccc}
\toprule
\textbf{Model} & \textbf{Accuracy} \\
\midrule
IPVR (OPT-66B) & 48.6 \\
ViLBERT & 49.1 \\
\midrule
$\text{MM-CoT}_{\rm{Base}}$ & 60.96 \\
$\textbf{ARES}_{\rm{Base}}$ (Ours) & \textbf{65.41}\\
\midrule
$\text{MM-CoT}_{\rm{Large}}$ & 65.68\\
$\textbf{ARES}_{\rm{Large}}$ (Ours) & \textbf{68.03}\\
\bottomrule
\end{tabular}
\end{adjustbox}
\centering
\caption{Results of ARES on A-OKVQA. We mainly compare different-sized MM-CoT baselines \citep{zhang2023multimodal}. We retrain the MM-CoTs and run the ARES pipeline on these models. We evaluate these models on the validation set because the test set is hidden.}
\label{table-a-okvqa}
\end{table}
\section{Related Work}
Chain-of-Thought (CoT) is a multi-step reasoning method for problem-solving that encourages LLMs to consider the intermediate reasoning steps. Zero-Shot-CoT \citep{kojima2023large} promotes CoT by using prompts such as \textit{"Let’s think step by step"} for LLMs. For Few-Shot-CoT \citep{zhang2022automatic, wei2023chainofthought}, a few examples with reasoning processes are provided, allowing the model to refer to these examples and understand how to perform CoT. \citet{wei2023chainofthought} reveal that this CoT technique positively impacts the performance of large models ($>100$B), but has minimal effect on smaller models. MM-CoT \citep{zhang2023multimodal} suggest that CoT is beneficial even for relatively small models, such as 200M, if the model that generates intermediate reasoning and the model that infers the answer are separated. We find that simply adding a LoRA adapter \citep{hu2021lora} to the reasoning model results in comparable performance. This framework enables the LoRA adapter to effectively utilize all features, from raw text to latent features, and generates answers 2x--14x faster than MM-CoT, which uses a separate inference model (See Table~\ref{time_comparison_scienceQA} and Table~\ref{time_comparison_aokvqa}). This speed advantage arises from the fact that our framework does not require a rationale as input, whereas the separate inference model framework must first generate the rationale before using it as input.

Reinforcement Learning from Human Feedback (RLHF) \citep{glaese2022improving, ouyang2022training} and AI Feedback (RLAIF) \citep{bai2022constitutional} align LLMs with user preferences. \citet{ouyang2022training} collects ranked feedback from human labelers and uses this feedback to perform Reinforcement Learning (RL). Constitutional AI (CAI) \citep{bai2022constitutional} collects ranked AI feedback rather than costly human feedback and handles harmfulness with RL. Both approaches learn outcome-supervised reward models (ORM) using ranking-based feedback. \citet{lightman2023lets}, instead, propose a process-supervised reward model (PRM) that leverages sentence-level feedback for CoT.  \citet{lightman2023lets, luo2024improve} evaluate each trained ORM and PRM with searching algorithms such as best-of-$N$ or Monte Carlo Tree Search (MCTS) by selecting the highest-scored solution, demonstrating that the PRM-selected solution outperforms the ORM-selected one. \citet{wang2024mathshepherd} perform RL using PRM, providing heuristic sentence-level scores for math problems that are simple to grade. As an LLM is trained with RL and starts generating outputs different from the original distribution, these reward models would not correctly provide rewards \citep{pitis2023failure, byun2024symmetric}.  Instead of training a reward model for a more general task, we perform RL by requesting sentence-level rewards from advanced AI models such as \verb|GPT-4|.
\begin{table}[t]
\centering
\begin{adjustbox}{max width=\textwidth}
\begin{tabular}{lcccccccccc}
\toprule
\textbf{Model} & \textbf{Accuracy} \\
\midrule
$\text{MM-CoT}_{\rm{Base}}$ & 85.95 \\
% \midrule
$\text{1st RL}$ & 86.70 \\
$\text{1st ARES}$ & 87.81 \\
% \midrule
$\text{1st ARES \&} \text{ 2nd RL}$ & 87.88 \\
$\text{2nd ARES}$ & 88.38 \\
\bottomrule
\end{tabular}
\end{adjustbox}
\centering
\caption{Ablation study: The accuracy gradually improves as each stage of ARES is added.}
\label{table-ablation}
\end{table}
\section{Conclusion}
We propose a hybrid algorithm, ARES, which \textbf{A}lternates \textbf{RE}inforcement Learning (RL) and \textbf{S}upervised Fine-Tuning (SFT) to enhance multi-modal rationale reasoning for ScienceQA and A-OKVQA. ARES leverages two types of feedback: $1)$ ARES requests a score from a Teacher (we used \verb|Claude 3 Haiku|) for sentence-level nuanced feedback and proceeds with RL. $2)$ ARES requests the Teacher to correct rationale chains after RL, stabilizing the RL fine-tuned model with SFT. ARES is designed to aid the RL procedure without massive hyperparameter tuning while properly reflecting the desired changes from RL. We evaluate the improvement in rationale reasoning produced by ARES compared to baselines using \verb|GPT-4o|, and assess how much the improved rationale chains enhance inference accuracy for the two multi-modal tasks. We hope our work inspires further research on utilizing various types of AI feedback.
\section*{Limitations}
Although we address general multi-modal rationale models beyond mathematical problems, receiving feedback from AI models still needs to be more reliable for more complex tasks such as graduate-level math or expert-level knowledge. For instance, some A-OKVQA problems even contain challenging questions requiring external knowledge beyond the image alone. This challenge highlights the necessity for future research to develop methods that can effectively incorporate external knowledge sources into the model. Additionally, if the model is not publicly available for free, using the API incurs costs, and there are daily usage limits.

% The A-OKVQA dataset includes challenging questions paired with rationales that require external knowledge beyond the image alone, which presents a limitation for our models relying solely on visual information. This limitation highlights the necessity for future research to develop and integrate methods that can effectively incorporate external knowledge sources into the model.

%Potential approaches include integrating advanced retrieval-based systems, improving the model's ability to access and utilize external databases, or enhancing its training with more comprehensive datasets that include relevant external information.

%To provide sentence-level reward feedback, we need to identify tokens representing each sentence's end, such as period marks, question marks, and exclamation marks. Additionally, ScienceQA often adds 'n' after the period mark due to newline characters in the dataset, and '\' is ignored. These exceptions vary for each dataset and require careful exception handling to distinguish sentences in token space.

\section*{Acknowledgments}
We sincerely appreciate the generous support of computational resources provided by the Ohio Supercomputer Center \citep{OhioSupercomputerCenter1987}.

\bibliography{custom}

\appendix

\section{Prompts}
\label{sec:appendix}
%\section{Example Appendix}

\begin{table*}[t]
  \centering
  \vspace{0.5cm}
  \begin{tabular}{m{4cm} m{11cm}}
    \toprule
    \textbf{Feedback} & \textbf{Prompt Structure} \\ 
    \midrule
    Sentence-Level \newline Nuanced Feedback & \textbf{[Prompt when Image is provided]}\newline There exists a set comprising Image, Options, Hint, and Answer for a Question. The reasoning process used to deduce the answer is provided in JSON format. Fill in "xxx" with values ranging from 0.0 to 1.0, in increments of 0.1. The reasoning may include the starting point of thought, the process of elimination, or true statements, although these may not appear to be directly related to the answer at first glance. A value closer to 0.0 indicates a completely incorrect rationale, 0.5 indicates a neutral rationale such as the initial thought process or true statements that guide later guesses towards the answer, and a value closer to 1.0 denotes a correct or relevant rationale for the question. Please just fill the "xxx" parts and only return the JSON format. If a sentence is repetitive (appeared before), then give 0.0. \newline\newline
    Question: <Question>\newline
Options: <Choices>\newline
Hint: <Hint>\newline
Answer: <Answer>\newline\newline
\{\newline
"<Rationale 1>": xxx,\newline
"<Rationale 2>": xxx,\newline
"<Rationale 3>": xxx\newline
\}
\newline\newline
\textbf{[Prompt when no Image is provided]}\newline There exists a set comprising Options, Hint, and Answer for a Question. The reasoning process ... \textit{<same as the prompt when the Image is provided>}\\
    
    \bottomrule
  \end{tabular}
  \caption{Prompt structure for sentence-level nuanced feedback in a question-answering system. The table outlines the format for prompts when an image is provided and when no image is provided, detailing how to score the rationale for each sentence in terms of correctness and relevance.}
  \label{fig:sentence_prompt}
\end{table*}

\begin{table*}[t]
  \centering
  \vspace{0.5cm}
  \begin{tabular}{m{4cm} m{10cm}}
    \toprule
    \textbf{Evaluation} & \textbf{Prompt Structure} \\ 
    \midrule
    Win Rate& You are given two rationales options (A or B). Your job is to select the better rationale between A and B for solving the given problem with the given Image, Choices, Hint, and Answer. Please output only A or B. \newline\newline
    Question: <Question>\newline
Choices: <Choices>\newline
Hint: <Hint>\newline
Answer: <Answer>\newline\newline
OPTION A: <Rationales> \newline
OPTION B: <Rationales> \\
    \bottomrule
  \end{tabular}
  \caption{Prompt structure used for evaluating the win rate of generated rationales using GPT-4o.}
  \label{fig:win_rate_prompt}
\end{table*}

%%%% 

\subsection{Prompt for Sentence-Level Nuanced Feedback}
\label{appendix_score}
The prompt for obtaining sentence-level nuanced feedback by \verb|Claude 3 Haiku| is illustrated in Table~\ref{fig:sentence_prompt}. Each reasoning sentence is assigned a value between $0.0$ and $1.0$. 
\begin{itemize}
    \item Values close to $0.0$ indicate completely incorrect rationales. 
    \item A value of $0.5$ represents a neutral rationale, such as an initial thought process or true statements that aid in guiding guesses towards the correct answer. 
    \item Values close to $1.0$ denote a correct or highly relevant rationale. 
\end{itemize}
These scores enable our model to discern the direction of changes through Reinforcement Learning (RL), reflecting the extent to which a sentence aids in resolving the problem.

\subsection{Prompt for Correction Feedback}
\label{appendix_correction}
Due to the challenges mentioned in Section~\ref{sec:RLchallenges}, we adopt the correction feedback approach. The following are the specific instructions for obtaining correction feedback using \verb|Claude 3 Haiku|. We have established the following seven rules for obtaining correction feedback using \verb|Claude 3 Haiku|: The prompt is presented as a Table~\ref{fig:correction_prompt}.

\subsection{Prompt for Win Rate Evaluation}
\label{appendix_win_rate}
We prompt \verb|GPT-4o (2024-05-13)| to choose which generated rationale is better for solving the question because we don't have gold rationales. 
Given two generated rationales (e.g., $\text{MM-CoT}_{\rm{Base}}$ and $\text{ARES}_{\rm{Base}}$), we ask \verb|GPT-4o|: "You are given two rationale options (A or B). Your job is to select the better rationale between A and B for solving the given problem with the given image, choices, hint, and answer. Please output only 'A' or 'B'." (see Table~\ref{fig:win_rate_prompt}). \citet{yu2023promptbased} find that ChatGPT is skewed towards choosing option A, so we randomly swap options A and B for each evaluation to avoid bias.

\section{Difficulties with External Knowledge in A-OKVQA}
\label{appendix_vqa_challenge}
The A-OKVQA dataset includes challenging questions paired with rationales that demand knowledge beyond the information available in the image. These questions cannot be answered simply by querying a knowledge base, as they require a deeper understanding and integration of external knowledge. 

Our model faces difficulties with problems that cannot be resolved using only the information from the image. While our approach is designed to improve rationales by addressing grammatical errors and incomplete or incorrect statements, it struggles with questions that necessitate external knowledge. %Consequently, our model has limitations in effectively handling such questions in the AOKVQA dataset. 

Table~\ref{fig:aokvqa-challenge1} illustrates an example where the question requires knowledge about the typical PSI range for bicycle tires. This information is not visually apparent from the image of the bicycle alone. To answer this question correctly, one needs external knowledge about standard bicycle maintenance practices and the recommended PSI ranges for different types of bicycle tires. This highlights a challenge for our model, as it must provide correct rationales and answers without access to such external knowledge, relying solely on the image and its internal knowledge base.

\begin{figure}[t]
  \includegraphics[width=\columnwidth]{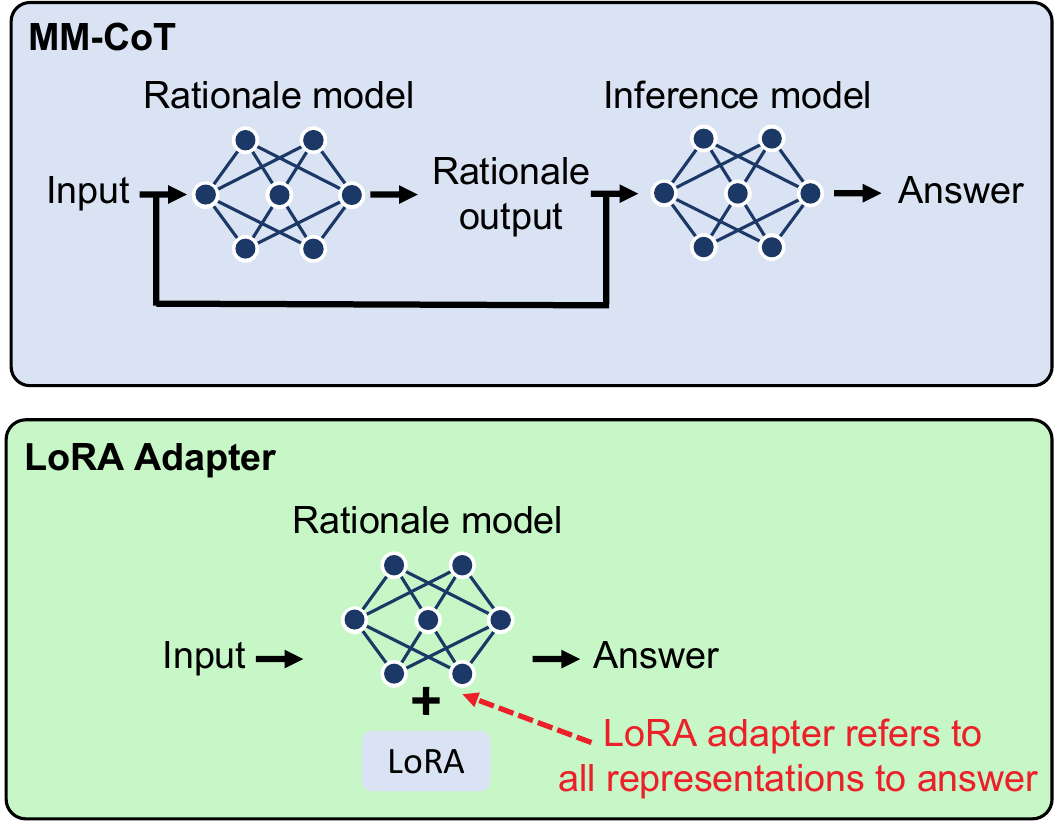}
   \caption{MM-CoT \citep{zhang2023multimodal} uses two same size separate models for reasoning and inference. We replace the inference model with a LoRA adapter (only $1/10$ the weights) added to the rationale model.}
  \label{fig:lora}
\end{figure}

In the second example (Table \ref{fig:aokvqa-challenge2}), understanding what the first two numbers of an identification tag denote requires specific external knowledge about livestock tagging systems. Such tagging systems often use coded information, where the numbers might represent the birth month, year, or other identification details. This information is not visually apparent from the image and requires familiarity with agricultural practices or livestock management, illustrating the challenge for our model in providing correct rationales and answers without external knowledge.

%For future work, addressing these challenges is essential for improving the performance and robustness of question-answering systems in domains requiring extensive external knowledge.

%The examples discussed highlight a limitation of our model: its reliance on internal knowledge and visual information from images alone, without the ability to integrate external knowledge that is often crucial for accurate reasoning and answering. This limitation underscores the need for future work to focus on methods for incorporating external knowledge sources into the model. Potential approaches include integrating advanced retrieval-based systems, improving the model's ability to access and utilize external databases, or enhancing its training with more comprehensive datasets that include relevant external information. Addressing these challenges will be essential for improving the performance and robustness of question-answering systems in domains requiring extensive external knowledge.

\section{Training Details}
\label{appendix_training_details}
ScienceQA has $21$K multi-modal problems, with $12$K for training, $4$K for validation, and $4$K for testing. It also includes various difficulty levels from elementary to high school, covering domains like natural science, language science, and social science. In addition, we conduct experiments on A-OKVQA \citep{schwenk2022aokvqa}, a knowledge-based multi-modal benchmark with a diverse set of 25K questions A-OKVQA includes 25K questions ($17$K for training, $1$K for validation, and $6$K for testing).

We adapt the same T5 encoder-decoder architecture \citep{raffel2023exploring} under Base and Large settings, following \citep{zhang2023multimodal}, and initialized with \text{Flan-Alpaca} \citep{chia2023flanalpaca}. 

\subsection{Reinforcement Learning}
For the $\text{ARES}_{\rm{Base}}$ and $\text{ARES}_{\rm{Large}}$ training on the ScienceQA and A-OKVQA dataset, we employ the following settings: \\

\noindent\textbf{Common Settings:} We use top-$k$ sampling with $k=50$ and sample 4 actions. The initial coefficient for the Kullback-Leibler (KL) divergence is set to $0.0001$. The range for clipping the probability ratios in PPO is $0.2$. The discount factor is set to $1.0$. Token length is constrained to $512$. We train the model using 4 NVIDIA A100 80GB GPUs. \\

\noindent\textbf{$\text{ARES}_{\rm{Base}}$ Specific Settings:} We use a learning rate of $2e{-5}$ and $10$ epochs for PPO with a batch size of $8$. Advantage normalization is applied for $\text{ARES}_{\rm{Base}}$, and gradient accumulation steps are set to $8$. \\

\noindent\textbf{$\text{ARES}_{\rm{Large}}$ Specific Settings:} The learning rate for $\text{Flan-Alpaca}_{\rm{Large}}$ is $2e{-5}$ with 5 epochs for PPO and a batch size of $2$ for both tasks. Advantage normalization is not used and gradient accumulation steps are set to $16$. \\

\subsection{Supervised Fine-Tuning}
\label{appendix_sft} 
We use a batch size of 8 and train for 20 epochs with a learning rate of $8e{-5}$ for $\text{ARES}_{\rm{Base}}$, following \citep{zhang2023multimodal}. For $\text{ARES}_{\rm{Large}}$, we use a batch size of $2$ and train for $50$ epochs with a learning rate of $5e{-5}$. The output length is set to $64$ tokens. Training for $\text{ARES}_{\rm{Base}}$ utilizes 1 A100 GPU, while training for $\text{ARES}_{\rm{Large}}$ utilizes 4 A100 GPUs. In the MM-CoT paper \citep{zhang2023multimodal}, because the final\_eval setting was not consistent, we retrained the base model with final\_eval=true and the large model with final\_eval=false for consistency. \\

\noindent\textbf{Token Cleanup:} 
In order to collect the corrected dataset, we need to identify tokens representing the end of each sentence, such as periods, question marks, and exclamation marks. In the ScienceQA dataset, a newline character often follows the `n' being added after it. To reduce the burden of feedback, we simply hard-code the removal of repetitive sentences before adding the generated rationale to the prompt. We remove this `n' and also ignore the backslash (\textbackslash) character. For overlapping sentences, we placed each rationale of a problem into a list. If a rationale sentence was already in the list, we did not include it again during this preprocessing step.

\subsection{LoRA Adapter Training}
\label{lora_adapter_traing}
MM-CoT utilizes two identically sized models for reasoning and inference tasks. In our approach, we replace the inference model with a LoRA adapter (Figure~\ref{fig:lora}), which is added to the rationale model and consists of only one-tenth of the weights.

For LoRA adapter training for ScienceQA and A-OKVQA, we use a LoRA rank of $r=64$, a LoRA $\alpha=128$, and a LoRA dropout rate of $0.05$. The learning rate is set to $8e{-5}$ for both $\text{ARES}_{\rm{Base}}$ and $\text{ARES}_{\rm{Large}}$. The batch size is $16$ for $\text{ARES}_{\rm{Base}}$ and $4$ for $\text{ARES}_{\rm{Large}}$ on the ScienceQA dataset. For the A-OKVQA dataset, the batch size is $4$ for both $\text{ARES}_{\rm{Base}}$ and $\text{ARES}_{\rm{Large}}$.

\subsection{Time Comparison between LoRA Adapter and Inference Model}
\label{time_comparison}
Rather than introducing a separate model for inference, we achieve comparable performance by adding the LoRA adapter to the rationale model, while simultaneously obtaining a 2x--14x speedup in inference (ARES) compared to MM-CoT, which introduces a separate inference model. We provide the time comparison in Table~\ref{time_comparison_scienceQA} and Table~\ref{time_comparison_aokvqa}.

This speed gap is mainly due to the fact that the separate inference model requires rationale generation before the inference procedure. However, the LoRA adapter directly refers to the rationale model's latent features to derive answers, eliminating the need to generate the rationale first.

\begin{table}[t]
\centering
\begin{adjustbox}{max width=\columnwidth}
\begin{tabular}{lccc}
\hline
\textbf{Model} & \textbf{Rationale} & \textbf{Inference} & \textbf{Total} \\ \hline
$\text{MM-CoT}_{\rm{Base}}$   & 1h 32m  & 7m  & 1h 39m \\ \hline
$\text{ARES}_{\rm{Base}}$     & -       & 8m  & 8m     \\ \hline
$\text{MM-CoT}_{\rm{Large}}$ & 5h 23m  & 12m & 5h 35m \\ \hline
$\text{ARES}_{\rm{Large}}$   & -       & 24m & 24m    \\ \hline
\end{tabular}
\end{adjustbox}
\caption{Time Comparison between MM-CoT and ARES models for ScienceQA test set}
\label{time_comparison_scienceQA}
\end{table}

\begin{table}[t]
\centering
\begin{adjustbox}{max width=\columnwidth}
\begin{tabular}{lccc}
\hline
\textbf{Model} & \textbf{Rationale} & \textbf{Inference} & \textbf{Total} \\ \hline
$\text{MM-CoT}_{\rm{Base}}$   & 6m  & 2m  & 8m \\ \hline
$\text{ARES}_{\rm{Base}}$     & -       & 3m  & 3m     \\ \hline
$\text{MM-CoT}_{\rm{Large}}$ & 16m  & 3m & 19m \\ \hline
$\text{ARES}_{\rm{Large}}$   & -       & 6m & 6m    \\ \hline
\end{tabular}
\end{adjustbox}
\caption{Time Comparison between MM-CoT and ARES models for A-OKVAQ test set}
\label{time_comparison_aokvqa}
\end{table}

\section{Comparison of Generated Rationales}
\label{appendix_comparision}
As mentioned in Section~\ref{sl_stage} and Section~\ref{ex_results_rl}, because RL increases the probability of sentences receiving positive rewards and reduces the probability of sentences receiving negative rewards, the trained model often exhibits specific phenomena. It tends to generate repetitive and incomplete sentences (Table~\ref{table-repeat} and Table~\ref{table-incompete}). Before the RL steps, the model couldn't produce rationales, but after RL steps, it starts generating meaningful rationale reasoning (Table~\ref{table-empty-filling}). Furthermore, it begins to generate reasons why other options are not the answer (Table~\ref{table-process-elimination}).

As illustrated in Table~\ref{fig:good_example}, we compare the solutions from the ScienceQA original dataset, the rationales generated by the baseline model ($\textbf{MM-CoT}_{\rm{Base}}$), the rationales from the baseline model with correction feedback applied, and the rationales generated by our model ($\textbf{ARES}_{\rm{Base}}$). The first example, "Which property do these three objects have in common?" illustrates that the baseline model generates incorrect rationales such as "The lemon is not (yellow)" and "All three objects are rough. The property that all three objects have in common is rough." However, when we apply correction feedback to the rationales generated by the baseline model and compare it to our proposed method, we see that our approach generates correct rationales that include the correct answer and provide explanations on why other options are not the answer. The second example also shows that our method improves rationale reasoning.

\begin{table*}[t]
  \centering
  \vspace{0.5cm}
  \begin{tabular}{m{4cm} m{11cm}}
    \toprule
    \textbf{Feedback} & \textbf{Prompt Structure} \\ 
    \midrule
    Correction Feedback & \textbf{[Prompt when Image is provided]}\newline Your task involves reviewing a set that includes an Image, Options, Hint, Answer, and Rationales for a Question. Please follow below 7 rules.\newline1. Preserve any correct original rationales based on the given answer by incorporating them into the final rationale without making any alterations.\newline
    2. Preserve any original rationales that represent the starting point of thought.\newline
    3. Correct any grammatical errors or incomplete rationales based on the given information without your knowledge.\newline
    4. If there are incorrect rationales based on the given answer, please correct them without removing them based on the given information.\newline
    5. Please take into account the content of the options, hint, and answer when doing this task.\newline
    6. Fill the corrected rationales inside the \{\} in the final\_rationale according to the given format below, without any additional explanation.\newline
    7. Return only the entire set of Rationales within curly braces (\{\}) below with the filled one in the step 6. 
 \newline\newline
    Question: <Question>\newline
Options: <Choices>\newline
Hint: <Hint>\newline
Answer: <Answer>\newline\newline
Rationales:\newline
\{\newline
original\_rationale:\{<Rationales>\}\newline
final\_rationale:\{\}\newline
\}\newline\newline
\textbf{[Prompt when no Image is provided]}\newline
Your task involves reviewing a set that includes Options, Hint, Answer, and Rationales for a Question. ... \textit{<same as the prompt when the image is provided>}\\
    \bottomrule
  \end{tabular}
  \caption{Prompt structure for correction feedback. The table details the rules and format for reviewing and correcting rationales when an image is provided and when no image is provided. Each set includes a question, options, hint, answer, and rationales, with specific instructions on preserving, correcting, and formatting the rationales.}
  \label{fig:correction_prompt}
\end{table*}

\begin{table*}[t]
  \centering
  \vspace{0.5cm}
  \begin{tabular}{>{\bfseries}m{14cm}}
    \toprule
    A Challenging Example from A-OKVQA  \\
    \midrule
  \end{tabular}
  \begin{tabular}{>{\bfseries}m{2cm} m{11cm}}
    Image & \includegraphics[width=0.8\linewidth]{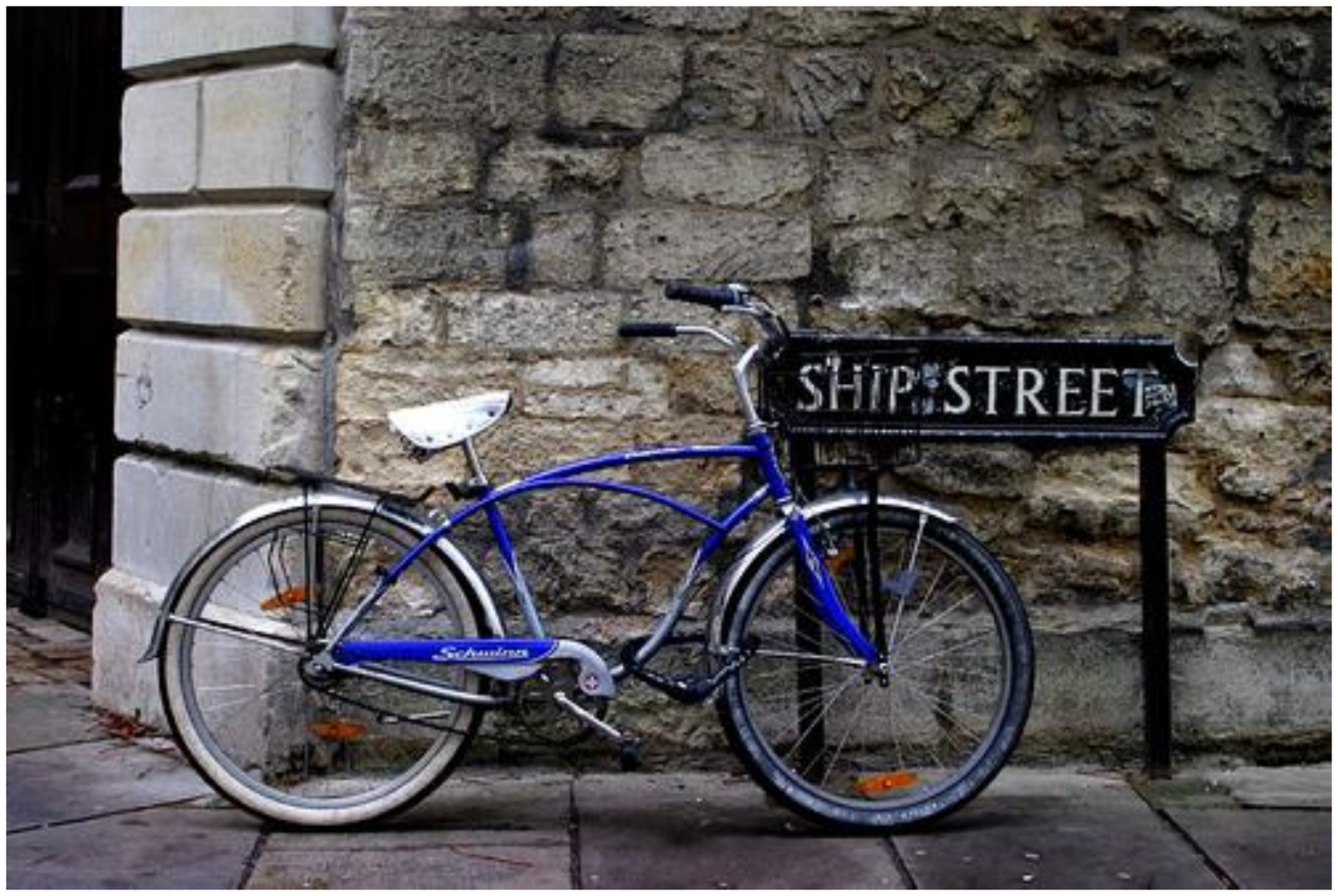} \\
    \midrule
    Question & What is the correct PSI for a bicycle tire? \\
    \midrule
    Choices & 100-120 psi, 80-130 psi, 40-90 psi, 50-80 psi \\
    \bottomrule
  \end{tabular}
   \caption{A challenging example question from the AOKVQA dataset. The question asks for the correct PSI for a bicycle tire, which requires external knowledge beyond what is depicted in the image. The choices provided represent typical PSI ranges for different types of bicycle tires.}
  \label{fig:aokvqa-challenge1}
\end{table*}

\begin{table*}[t]
  \centering
  \vspace{0.5cm}
  \begin{tabular}{>{\bfseries}m{14cm}}
    \toprule
    A Challenging Example from A-OKVQA \\
    \midrule
  \end{tabular}
  \begin{tabular}{>{\bfseries}m{2cm} m{11cm}}
    Image & \includegraphics[width=0.8\linewidth]{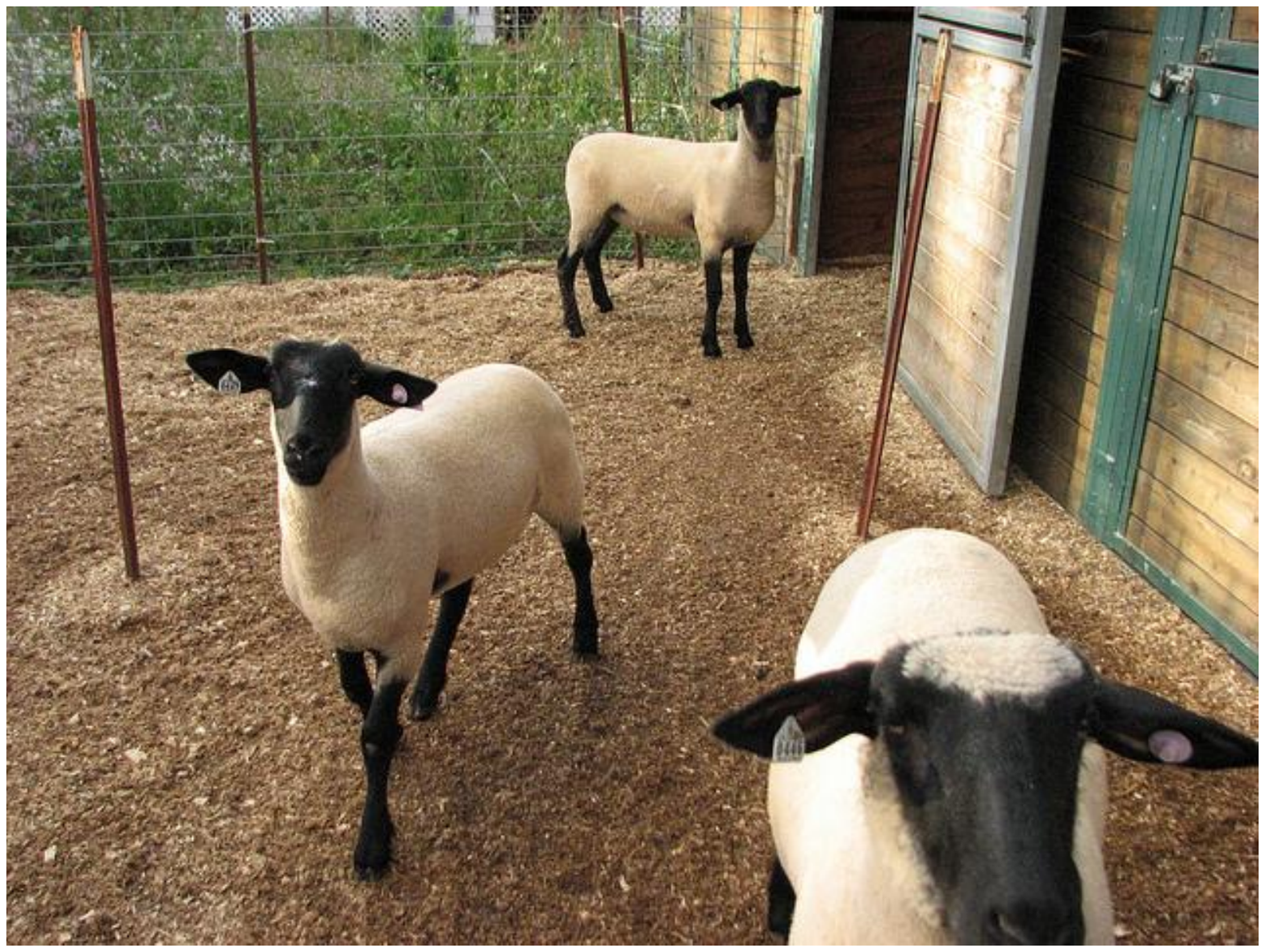}\\
    \midrule
    Question & The first two numbers of the identification tag denote what? \\
    \midrule
    Choices & birth month, own number, birth year, birth date \\
    \bottomrule
  \end{tabular}
  \caption{A challenging example question from the AOKVQA dataset. The question asks for the meaning of the first two numbers on an identification tag, which requires external knowledge about livestock tagging systems. The choices provided represent typical information encoded in such tags.}
  \label{fig:aokvqa-challenge2}
\end{table*}

%The SFT stage, with correction feedback, reflects what RL wants to achieve and appropriately guides it.
\begin{table*}[ht]
\centering
\begin{tabular}{p{4cm} p{11cm}}
\toprule
\multicolumn{2}{c}{\textbf{Example of the Repetitive Sentences After the RL Stage}} \\
\midrule
\multicolumn{2}{l}{\textbf{Problem:} Which tense does the sentence use? "Mona will print her name with care."} \\
\multicolumn{2}{l}{\textbf{Choices:} [present tense, future tense, past tense]} \\
\multicolumn{2}{l}{\textbf{Answer:} future tense} \\
\midrule
Before RL & \textbf{[Rationale reasoning]}\\
& The sentence is in future tense. You can tell because it uses will before the main verb, print. The verb tells you about something that is going to happen.\\
\midrule
After RL & \textbf{[Rationale reasoning]} \\
& The sentence is in future tense. You can tell because it uses will before the main verb, print. The verb tells you about something that is going to happen. \textcolor{red}{The verb tells you about something that is going to happen. The verb tells you about something that is going to happen. The verb ends in -s and tells you about something that is going to happen...} \\
\bottomrule
\end{tabular}
\caption{Example of the repetitive sentences after the RL stage.}
\label{table-repeat}
\end{table*}

\begin{table*}[ht]
\centering
\begin{tabular}{p{4cm} p{11cm}}
\toprule
\multicolumn{2}{c}{\textbf{Example of the Incomplete 
 Sentence After the RL Stage}} \\
\midrule
\multicolumn{2}{l}{\textbf{Problem:} What is the volume of a large soup pot?} \\
\multicolumn{2}{l}{\textbf{Choices:} [7 liters, 7 milliliters]} \\
\multicolumn{2}{l}{\textbf{Answer:} 7 liters} \\
\midrule
Before RL & \textbf{[Rationale reasoning]}\\
& The better estimate for the volume of a large soup pot is 7 liters. 7 milliliters is too little.\\
\midrule
After RL & \textbf{[Rationale reasoning]} \\
& The better estimate for the volume of a large soup pot \textcolor{red}{is-}\\
\bottomrule
\end{tabular}
\caption{Example of the incomplete sentence after the RL stage.}
\label{table-incompete}
\end{table*}

\begin{table*}[ht]
\centering
\begin{tabular}{p{4cm} p{11cm}}
\toprule
\multicolumn{2}{c}{\textbf{Changed Empty Rationale Reasoning Through RL}} \\
\midrule
\multicolumn{2}{l}{\textbf{Problem:} Select the solid.} \\
\multicolumn{2}{l}{\textbf{Choices:} [rain, water in a fishbowl, hammer]} \\
\multicolumn{2}{l}{\textbf{Answer:} hammer} \\
\midrule
Before RL & \textbf{[Rationale reasoning]}\\
& <empty> \\
\midrule
After RL & \textbf{[Rationale reasoning]} \\
& \textcolor{blue}{Rain is a liquid. A liquid takes the shape of any container it is in. If you put rainwater into a bucket, the rainwater will take the shape of the bucket. But the rainwater will still take up the same amount of space. The water in a fishbowl is a liquid. A liquid takes the shape of any container it is in. If you pour water from a fishbowl into a different container, the water will take the shape of that container. But the water will still take up the same amount of space. A hammer is a solid. A solid has a size and shape of its own. Many hammers are made of iron and wood.The water in a fishbowl is a liquid. A liquid takes the shape of any container it is in. If you put water from a fishbowl into a different container, the water will take the shape of that container. But the water will still take up the same amount of space. A rain is a liquid. A liquid takes the shape of any container it is in. If you put rainwater into a bucket, the rainwater will take the shape of the bucket. But the rainwater will still take up the same amount of space. A hammer is a solid. A solid has a size and shape of its own. Many hammers are made of iron and wood. Both iron and wood are solids.} \\
\bottomrule
\end{tabular}
\caption{Example of an empty rationale reasoning that starts generating rationale reasoning after the RL stage.}
\label{table-empty-filling}
\end{table*}

\begin{table*}[ht]
\centering
\begin{tabular}{p{4cm} p{11cm}}
\toprule
\multicolumn{2}{c}{\textbf{Example of the Process of Elimination After the RL Stage}} \\
\midrule
\multicolumn{2}{l}{\textbf{Problem:} Which is a compound sentence?} \\
\multicolumn{2}{l}{\textbf{Choices:} ["Last night we heard an owl outside in the oak tree.",} \\
\multicolumn{2}{l}{\hspace{1.5cm}"I ate all of my dinner, so I can have some pie."]}\\
\multicolumn{2}{l}{\textbf{Answer:} "I ate all of my dinner, so I can have some pie."} \\
\midrule
Before RL & \textbf{[Rationale reasoning]}\\
& The first sentence is the compound sentence. It is made up of two simple sentences joined by a comma and the conjunction so. I ate all of my dinner, so I can have some pie. \\
\midrule
After RL & \textbf{[Rationale reasoning]} \\
& The first sentence is the compound sentence. It is made up of two simple sentences joined by a comma and the conjunction so. I ate all of my dinner, so I can have some pie. \textcolor{blue}{The second sentence does not have a comma, so it is not a compound sentence. Last night we heard an owl outside in the oak tree.} \\
\bottomrule
\end{tabular}
\caption{Example of the process of elimination rationale reasoning after the RL stage.}
\label{table-process-elimination}
\end{table*}

\begin{table*}[t]
  \centering
\begin{tabular}{p{4cm} p{11cm}}
  \end{tabular}
  \begin{tabular}{>{\bfseries}m{2cm} m{12cm}}
  \toprule
    \multicolumn{2}{c}{\textbf{Comparison of Corrected Rationales After RL}}\\
    \midrule
    Image & \includegraphics[width=0.4\linewidth]{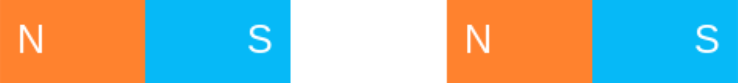}\\
    \midrule
    Question & Will these magnets attract or repel each other? \\
    \midrule
    Choices & repel, \hl{attract}\\
    % \midrule
    % \textbf{Original \newline Solution} & Will these magnets attract or repel? To find out, look at which poles are closest to each other. The south pole of one magnet is closest to the north pole of the other magnet. Poles that are different attract. So, these magnets will attract each other.\\
    \midrule
    \textbf{After RL} & Will these magnets attract or repel? To find out, look at which poles are closest to each other. The north pole of one magnet is closest to the south pole of the other magnet. Poles that are the same repel. So, these magnets will repel each other. The picture shows two magnets that are the same shape and the same size. Both magnets are the same size. So, these magnets will repel each other. \\ 
    \midrule
    \textbf{Corrected Rationale} & The magnets will attract each other because the north pole of one magnet is closest to the south pole of the other magnet. Opposite poles of magnets attract each other in this diagram. Therefore, the correct answer is that the magnets will attract. \\
    \bottomrule
    \toprule
    Question & How long is a pen? \\
    \midrule
    Choices & 20 kilometers, \hl{20 centimeters} \\
    % \midrule
    % \textbf{Original \newline Solution} & The better estimate for the length of a pen is 20 centimeters. 20 kilometers is too long. \\
    \midrule
    \textbf{After RL} & The better estimate for the length of a pen is 20 centimeters. 20 kilometers is too long. 20 centimeters is too short. 20 kilometers is too long. \\
    \midrule
    \textbf{Corrected Rationale} & The better estimate for the length of a pen is 20 centimeters. 20 kilometers is an unreasonable estimate, as it is far too long for a typical pen. 20 centimeters is a more accurate and reasonable estimate for the length of a standard pen. \\
    \bottomrule
    \toprule
    Image & \includegraphics[width=0.4\linewidth]{images/table17_2.pdf}\\
    \midrule
    Question & What is the capital of South Carolina?\\
    \midrule
    Choices & \hl{Columbia}, Montgomery, Charleston, Harrisburg \\
    % \midrule
    % \textbf{Original \newline Solution} & Columbia is the capital of South Carolina. \\
    \midrule
    \textbf{After RL} & Columbia is the capital of South Carolina.nThe city of Columbia is Charleston, South Carolina. Charleston is the smallest city in South Carolina.nHayeburg is the smallest city in South Carolina. Montgomery is the smallest city in South Carolina. But the city of Columbia is home to many cultural and historical landmarks. \\
    \midrule
    \textbf{Corrected Rationale} & Columbia is the correct answer as it is the capital city of South Carolina. It is the largest city in the state and serves as the seat of government. The other options, Montgomery (the capital of Alabama), Charleston (a city in South Carolina), and Harrisburg (the capital of Pennsylvania), are not the capital of South Carolina.\\
    \bottomrule
  \end{tabular}
  \caption{Examples of how the correction stage corrects mistakes after the RL stage.}
  \label{fig:correction_example}
\end{table*}

\begin{table*}[t]
  \centering
\begin{tabular}{p{4cm} p{11cm}}
  \end{tabular}
  \begin{tabular}{>{\bfseries}m{2cm} m{12cm}}
  \toprule
    \multicolumn{2}{c}{\textbf{Comparison of Generated Rationales}}\\
    \midrule
    Image & \includegraphics[width=0.5\linewidth]{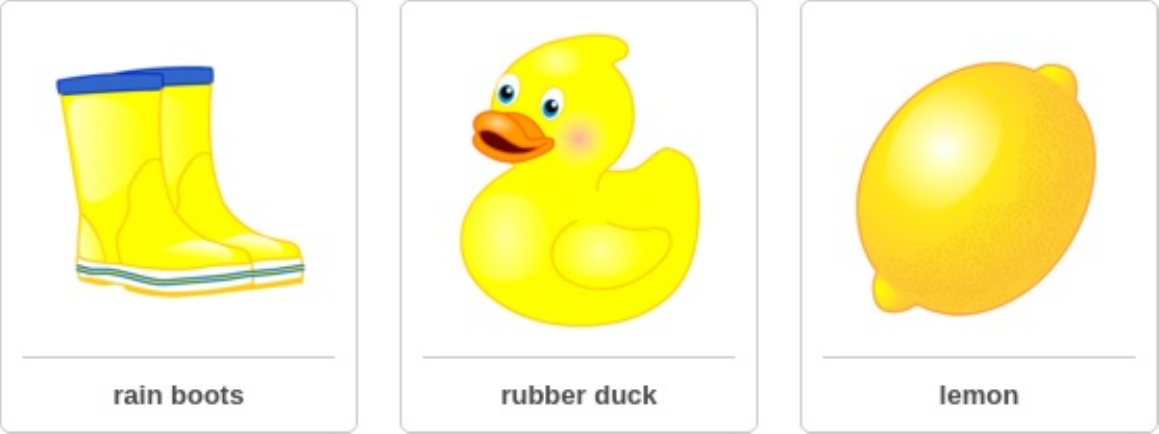}\\
    \midrule
    Question & Which property do these three objects have in common? \\
    \midrule
    Choices & rough, \hl{yellow}, sticky\\
    \midrule
    \textbf{Original \newline Solution} & Look at each object. For each object, decide if it has that property. A rough object feels scratchy when you touch it. The rain boots and the rubber duck are not rough. A sticky object can attach or stick to other things. The rain boots and the rubber duck are not sticky. Yellow is a color. This color is yellow. All three objects are yellow. The property that all three objects have in common is yellow.\\
    \midrule
    $\textbf{MM-CoT}_{\rm{Base}}$ & Look at each object. For each object, decide if it has that property. A sticky object can attach or stick to other things. The rubber duck is not sticky. Yellow is a color. This color is yellow. The rubber duck and the rain boots are yellow, but the lemon is not. A rough object feels scratchy when you touch it. All three objects are rough. The property that all three objects have in common is rough. \\ 
    \midrule
    \textbf{Correction \newline feedback on $\textbf{MM-CoT}_{\rm{Base}}$} & The property that all three objects have in common is yellow. The rubber duck and the rain boots are both yellow, and the lemon is also yellow. Although the objects may have other properties like roughness, the common property among them is the color yellow. \\
    \midrule
    $\textbf{ARES}_{\rm{Base}}$ (Ours) &  The three objects shown - rubber duck, rain boots, and lemon - all share the common property of being yellow in color. The rubber duck is made of a smooth, non-sticky material, the rain boots are also a yellow-colored object, and the lemon is also a yellow-colored liquid. Therefore, the correct answer is that the common property among these three objects is that they are yellow. The other options, rough and sticky, do not apply to all three objects, as the rubber duck is not rough, and the rain boots are not sticky. \\
    \bottomrule
    \toprule
    Question & Which word would you find on a dictionary page with the following guide words? lever - loan \\
    \midrule
    Choices & lamb, \hl{liquid} \\
    \midrule
    \textbf{Original \newline Solution} & Put the words in alphabetical order. Since liquid is between the guide words lever - loan, it would be found on that page. \\
    \midrule
    $\textbf{MM-CoT}_{\rm{Base}}$ & Put the words in alphabetical order. Since lamb is between the guide words lever - loan, it would be found on that page.\\
    \midrule
    \textbf{Correction \newline feedback on $\textbf{MM-CoT}_{\rm{Base}}$} & The word "liquid" would be found on the dictionary page with the guide words "lever - loan" because it falls alphabetically between those two words. \\
    \midrule
    $\textbf{ARES}_{\rm{Base}}$ (Ours) &  The word "liquid" would be found on the dictionary page with the guide words "lever - loan" because it falls alphabetically between those two words. Therefore, the word "liquid" would be the word found on that dictionary page. \\
    \bottomrule
  \end{tabular}
  \caption{Comparison of generated rationales by baseline, baseline with correction feedback, and our model (ARES).}
  \label{fig:good_example}
\end{table*}

\end{document}